\documentclass[conference]{IEEEtran}
\usepackage{comment}
\IEEEoverridecommandlockouts
\usepackage{cite}
\usepackage{amsmath,amssymb,amsfonts}
\usepackage{url}
\usepackage{graphicx}
\usepackage{textcomp}
\usepackage{xcolor}
\usepackage{amsmath}
\usepackage{amssymb}
\usepackage{algorithm}
\usepackage{algorithm}
\usepackage{algpseudocode}

\usepackage{amsthm}
\usepackage{comment}
\theoremstyle{definition}
\newtheorem{definition}{Definition}[section]
\newtheorem{theorem}{Theorem}[section]
\newtheorem{lemma}{Lemma}[section]

\def\BibTeX{{\rm B\kern-.05em{\sc i\kern-.025em b}\kern-.08em
    T\kern-.1667em\lower.7ex\hbox{E}\kern-.125emX}}
\begin{document}

\title{Typed-Centrality Grounding for Agentic GraphRAG: A Symbolic Approach for Agentic Reasoning}

\author{
\IEEEauthorblockN{Rosario Napoli}
\IEEEauthorblockA{\textit{University of Messina}\\
Messina, Italy \\
0009-0006-2760-9889} 
\and
\IEEEauthorblockN{Antonio Celesti}
\IEEEauthorblockA{\textit{University of Messina}\\
Messina, Italy \\
0000-0001-9003-6194} 
\and
\IEEEauthorblockN{Massimo Villari}
\IEEEauthorblockA{\textit{University of Messina}\\
Messina, Italy \\
0000-0001-9457-0677} 
\and
\IEEEauthorblockN{Maria Fazio}
\IEEEauthorblockA{\textit{University of Messina}\\
Messina, Italy \\
0000-0003-3574-1848} 
}

\maketitle

\begin{abstract}
Knowledge Graphs (KGs) and Large Language Models (LLMs) represent two approaches for knowledge representation and reasoning over heterogeneous data. Recent research has explored their integration through Graph Retrieval-Augmented Generation (GraphRAG) systems, which use KGs to provide structured context to LLMs. However, existing GraphRAG approaches treat KGs as homogeneous graphs and rely on textual similarity between the input prompt and the context, rather than using it as a source of deterministic knowledge.

This work proposes an agent-based GraphRAG system that introduces a symbolic layer to integrate deterministic reasoning alongside text similarities. This approach is based on the definition of typed centralities for KGs, where centrality measures are computed separately for each specific KG's class. This information is used to construct a symbolic representation that is provided to the LLM as a deterministic source. The proposed system was mathematically validated and evaluated on a real-world POLE (Person, Object, Location, Event) dataset, comparing reasoning and information retrieval with and without the symbolic layer. The results indicate that the our approach leads to responses that are more consistent with the context, improving the generated outputs.
\end{abstract}


\section{Introduction}


In order to get value from large-scale heterogeneous data, research is increasingly moving toward knowledge-based approaches for structured representation and AI-based methodologies for data processing \cite{rosario1}. Knowledge Graphs (KGs) are graph-based data structures that represent different types of nodes and relationships using labels to describe their nature. They combine the principles of graph theory with the semantic meaning of labels to extract information from data \cite{b23}.

While KGs use a deterministic strategy, Large Language Models (LLMs) have emerged as an AI-based probabilistic approach. LLMs work through statistical distributions learned from sequences of tokens, linking the input prompt to the knowledge acquired during extensive pre-training \cite{a24}. Both approaches have shown great effectiveness in the analysis of heterogeneous data, achieving significant results in areas such as healthcare and finance \cite{rosario3}.

The recent literature aims to combine the deterministic rigor of KGs with the probabilistic approach of LLMs, with the aim of improving information analysis over data \cite{r9}. In this context, Graph Retrieval-Augmented Generation (GraphRAG) has emerged as one of the most effective hybrid mechanisms to link these two strategies, allowing LLM responses to be grounded in a structured context provided by the KG \cite{a26}. 

Despite the high capabilities demonstrated by GraphRAG systems in complex reasoning and information retrieval, they inherit many of the limitations of the two methodologies considered individually \cite{r10}. On the one hand, KGs model reality through deterministic centrality measures that treat the underlying structure as a homogeneous graph, producing values that do not accurately reflect the represented domain \cite{rosario2}. On the other hand,  LLMs' responses generation is only based on textual similarity between the provided input and the available knowledge, which in GraphRAG systems is represented by the labels on the graph. This leads to the production of semantically plausible but incorrect responses, a phenomenon commonly known as hallucinations \cite{a25}. 
Therefore, GraphRAG works as a retrieval-enhanced framework where KGs act as additional textual information to augment the LLM, providing supplementary context rather than enforcing a strictly deterministic model of reality.

To achieve this full integration, this work proposes a GraphRAG system that addresses  their respective limitations. To better model the context from the KG, we introduce the concept of typed centrality, where centrality measures are computed separately for each specific KG's class. These values are used to build the Symbolic Layer, an additional context that acts as a deterministic base for the GraphRAG system. This allows inference and information retrieval to rely not only on textual similarity but also on symbolic measures. As a result, LLM's knowledge is grounded in a more accurate context, since the generated responses  depend on both probabilistic and deterministic information.
This results in: stronger grounding, fewer hallucinations and improved answer explainability.

To validate the effectiveness of this approach, the system was mathematically validated and tested on a real POLE (Person, Object, Location, Event) dataset, conducting a series of experiments aimed at comparing reasoning capabilities with and without the Symbolic Layer. The results showed that our integration gave a deterministic anchor to the data, with answers that were more consistent with the given context.

The remainder of this paper is structured as follows.
Section \ref{sec:related_work} reviews the current state of the art and the inherent limitations of existing GraphRAG systems. The core of our contribution is detailed in Section \ref{sec:architecture}, which describes the proposed four-layer hierarchical architecture and the implementation of the Symbolic Layer. Section \ref{sec:model_validation} provides a formal validation of the framework. The technical background is defined in Section \ref{sec:background}, while the experimental setup and the evaluation are detailed across Sections \ref{sec:experimental_setup} and \ref{sec:experimental_eval}. We report and discuss the results in Section \ref{sec:results} before  concluding and outlining potential directions for future research in Section \ref{sec:conclusion}.

\section{Related Work}
\label{sec:related_work}
The evolution of information retrieval systems has recently shifted toward hybrid architectures, as the research community seeks more robust methodologies to ground generative AI in structured contexts \cite{r13}.
Early RAG approaches were introduced to expand LLMs' knowledge by retrieving external textual information to supplement what was learned during their initial training \cite{r5}. These techniques were further extended through the use of KGs, giving rise to GraphRAG systems that integrate structured knowledge into the LLM~\cite{a10}. The transition from standard RAG pipelines to GraphRAG has been facilitated by protocols such as MCP~\cite{a11}, which standardize interactions between LLMs and external tools, integrating knowledge retrieval and reasoning directly into the LLM’s decision-making loop.
However, most existing implementations retain the classical assumptions of standard RAG pipelines \cite{r12}. Specifically, KGs are traversed to extract textual information from nodes and relationships that are added in the context. Nonetheless, there remains a lack of consideration of the graph structure as a deterministic guide for the inference process \cite{r6}.

Within this setting, recent GraphRAG's benchmarks primarily evaluate the LLM’s ability to select and invoke the correct tools for a specific task~\cite{a12}. However, these criteria do not consider whether the information retrieved is actually consistent with the KG's semantics.
This limitation becomes particularly evident in distributed agentic systems, where decisions are made across multiple sequential reasoning steps. As shown in~\cite{a16}, a semantic misinterpretation during the early information selection process can propagate through the reasoning chain, leading to outputs that are procedurally valid but grounded in incorrect assumptions. 

Several works have attempted to partially address this limitation by introducing retrieval criteria based on graph's properties. For instance, meta-path–based methodologies restrict the search space using predefined sequences of node and relation types~\cite{a15}. These methods heavily rely on domain knowledge to define meaningful paths. As a consequence, they struggle to generalize without extensive manual intervention. Other approaches like post-hoc verification methods operate exclusively on final outputs but do not influence intermediate retrieval or planning decisions made during execution~\cite{a17}. So, they fail to prevent error propagation during the reasoning chain, making it impossible to get the system back on track once an early mistake occurs.
Other studies propose heterogeneous or type-aware node representations that preserve semantic information within embedding spaces~\cite{a21}. Nonetheless, these techniques are typically employed as isolated retrieval components and evaluated on supervised tasks such as link prediction or node classification, rather than being integrated into the iterative decision-making loop of an LLM agent.

Recent proposals such as \cite{r1} utilize graph neural networks (GNNs) to improve KG retrieval for multi-hop Questions-Answers (QA), but the paths are verbalized and fed to the LLM without embedding semantic constraints into agent reasoning. Similarly, iterative frameworks like KG-IRAG aim to integrate multi-stage retrieval for temporal reasoning, but they do not formalize KG schema information within the agent’s real-time decision loop \cite{r2}. Works such as GRIL \cite{r3} explore joint graph-LLM optimization, reducing dependence on entity annotations, but still treats retrieval and reasoning as separated processes. Other contributions like KG²RAG and K-RagRec \cite{r4} apply graph-guided retrieval to improve coherence and downstream tasks such as recommendation, but they similarly fall short of using deterministic guidance during reasoning.

Recent surveys highlight that even if LLM-based agents have the capacity to plan multi-step queries, their actual navigation refers only to global metrics or simple text similarity rather than deep semantic reasoning \cite{a12}. As a result, current GraphRAG systems  rely on embedding-based retrieval that treats graph's paths as unstructured text blocks, prioritizing semantic proximity in a latent vector space \cite{r8}. Some existing approaches attempt to construct contextual information by relying on global centrality measure but this causes a misinterpretation of the KG's modeled reality \cite{r7}. Specifically, these methods treat all nodes uniformly, losing the semantic diversity of the KG and leading to outputs that are coherent in language but inconsistent with the graph's organization.

\section{Proposed Architecture: Typed-Centrality Grounding for Agentic GraphRAG}
\label{sec:architecture}
To the best of our knowledge, there is currently no complete integration of KGs-based symbolic reasoning with LLM-based approaches. Addressing this gap requires a shift from passive retrieval toward real hybrid approaches. By using KG-based centrality measures for symbolic reasoning, the GraphRAG can actually have a source of deterministic knowledge rather than a simple text concatenation. This enables the combination of the deterministic structure of KGs with the probabilistic nature of LLMs, allowing the data retrieval process to be also grounded on both semantics and graph-modeled reality.

\begin{figure}[ht!]
\centering
\includegraphics[width=0.95\linewidth]{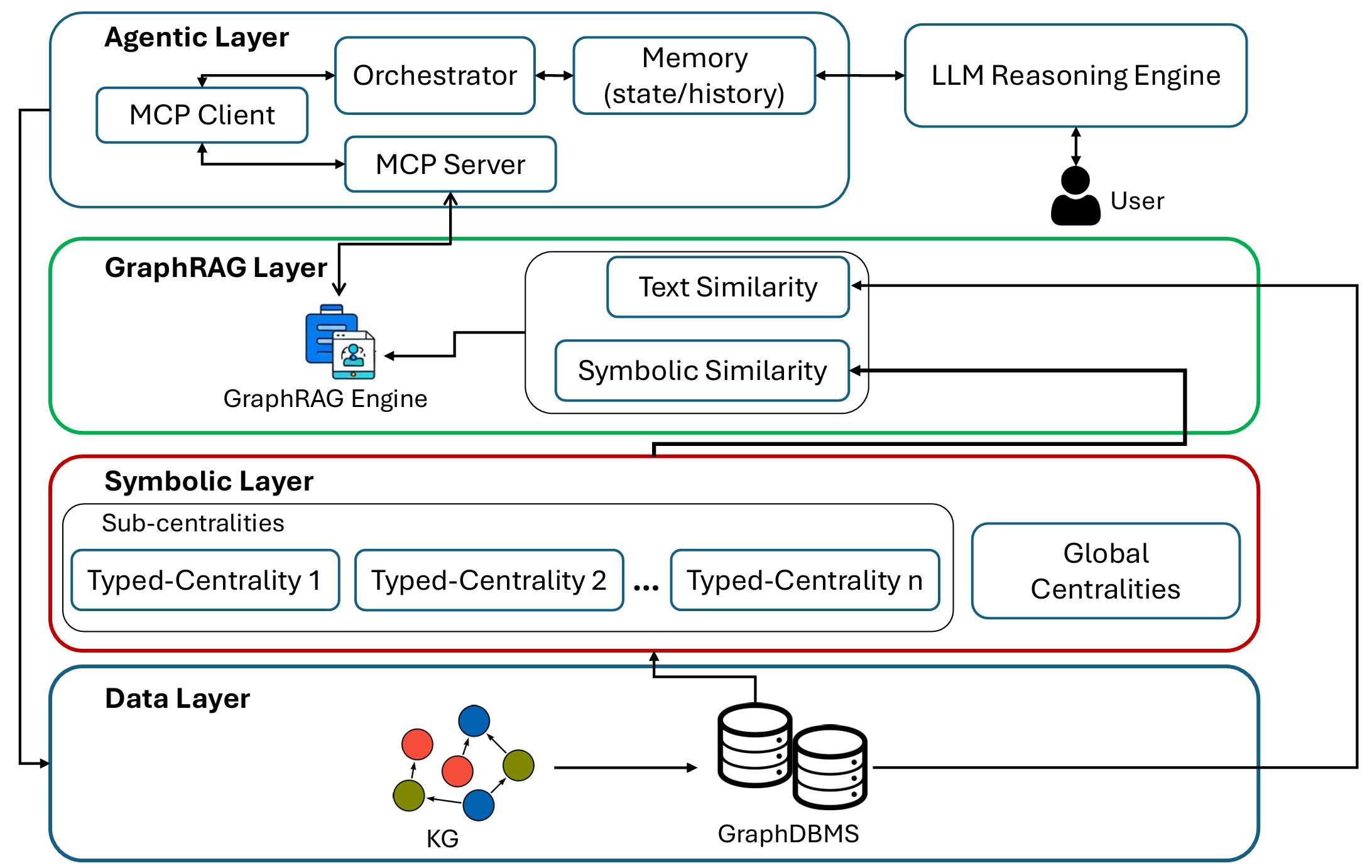}
\caption{Layered architecture of the Typed-Centrality Grounding framework.}\label{fig:tcg-architecture}
\end{figure}

To address the identified limitations, we introduce the Typed-Centrality Grounding for Agentic GraphRAG, a framework that treats KGs as deterministic signals for LLM's reasoning. This hybrid approach integrates the symbolic nature of KGs in GraphRAGs by redefining centrality measures as a deterministic background. Instead of applying heterogeneous metrics across the entire network, the framework represents centrality as a composition of class-specific measures calculated independently for each node type. This method allows for a coherent characterization of the modeled context by considering the different semantics within the graph. These type-specific metrics are used in the Symbolic Layer, an additional input which acts as a deterministic anchor to ground the GraphRAG inference process. The overall architecture is organized into four layers, spanning from data storage to the reasoning loop (Fig.~\ref{fig:tcg-architecture}).

\subsection{Data Layer}
The Data Layer provides the source of truth by storing the KG in a Graph Database Management System (DBMS). This layer represents the main context from which the LLM agent will perform its reasoning.

\subsection{Symbolic Layer}
The Symbolic Layer is the core component of the framework, responsible for computing typed centrality measures. Specifically, the layer divides the KG by nodes type, producing type-specific subgraphs. Standard centrality measures (e.g., PageRank and Betweenness) are then computed independently within each subgraph, resulting in Sub-Centralities Vectors that quantify the relative node importance over the KG. To calculate each typed-centrality, a topology-based mechanism classifies nodes according to local connectivity patterns, specifically:

\begin{enumerate}
\item Direct Connection (1-hop): entities with direct connections or self-loops are directly represented as subgraphs.
\item Structural Bridges (2-hop): For node types that lack self connections, the system identifies latent associations by treating the schema as a bipartite network. In particular, it uses shared bridge nodes at distance one to perform a one-mode projection between entities of the same type. This mechanism allows us to compute centrality even in the absence of explicit direct links, capturing the connectivity through intermediate common neighbors.
\item Isolated Entities: nodes without connections are excluded from the centrality computation.
\end{enumerate}

\subsection{GraphRAG Layer}
This layer integrates the structural signals from the Symbolic Layer with textual similarity. Subgraphs are analyzed according to prompt-based textual similarity and type-specific centrality.

\subsection{Agentic Layer: Reasoning and Orchestration}
The top layer implements an autonomous reasoning loop, specifically:

\begin{itemize}
\item LLM Reasoning Engine: interprets user input, determines high-level reasoning strategies and directs the retrieval process.
\item Orchestrator: coordinates execution, manages conversation state and interfaces with a Context Client.
\item Context Client and Server: the client translates orchestrator commands into protocol-specific requests, while the server exposes the GraphRAG Layer.
\end{itemize}

\section{Model Validation}
\label{sec:model_validation}

In this section, we formally demonstrate why global centrality metrics fail to capture the reality of KGs and we prove that our type-partitioned approach restores the necessary signal for accurate reasoning.

\begin{definition}[Global vs. Typed Graph]
Let $G = (V, E)$ be the global KG containing all nodes $V$ and edges $E$.
Let $G_{\tau} = (V_{\tau}, E_{\tau})$ be a type-specific subgraph, where $V_{\tau} \subset V$ contains only nodes of a specific entity type $\tau$, and $|V_{\tau}|$ represents the number of nodes in that type.
\end{definition}

\begin{definition}[Centrality Magnitude]
Let $C(v, G)$ be the centrality score of a node $v$ computed within graph $G$. Since centrality metrics represent a probability distribution, the sum of scores over the graph is normalized to 1:
\begin{equation}
    \sum_{v \in G} C(v, G) = 1
\end{equation}
\end{definition}

We first prove that mixing different entity types causes relevant signals to vanish.

\begin{lemma}[The Dilution Lemma]
\label{lemma:dilution}
In a global graph $G$, if the total number of nodes $|V|$ is significantly larger than the number of relevant target nodes $|V_{\tau}|$ (i.e., $|V| \gg |V_{\tau}|$), then the average centrality score of a target node approaches zero.
\end{lemma}

\begin{proof}
Consider the average centrality score $\mu$ in the global graph. Since the total probability is 1 and is distributed across $|V|$ nodes:
\begin{equation}
    \mu_{\text{global}} \approx \frac{1}{|V|}
\end{equation}
If the graph is dominated by irrelevant nodes, $|V|$ becomes very large. Consequently, the score for any specific relevant node $v \in V_{\tau}$ is suppressed by the massive denominator $|V|$.
\begin{equation}
    C(v, G) \propto \frac{1}{|V|} \approx 0
\end{equation}
This makes the relevant node statistically indistinguishable from noise.
\end{proof}

We now demonstrate how the Symbolic Layer resolves this limitation.

\begin{theorem}[Signal Restoration]
\label{theorem:restoration}
Computing centrality within the type-partitioned subgraph $G_{\tau}$ amplifies the signal of relevant nodes by a factor proportional to the ratio of irrelevant data removed.
\end{theorem}

\begin{proof}
When we project to the subgraph $G_{\tau}$, we remove all irrelevant nodes ($V - V_{\tau}$). The new domain size is $|V_{\tau}|$.
The average centrality score $\mu_{\tau}$ in this projection becomes:
\begin{equation}
    \mu_{\tau} \approx \frac{1}{|V_{\tau}|}
\end{equation}
We define the \textit{Signal Amplification Factor} $A$ as the ratio between the typed score and the global score:
\begin{equation}
    A = \frac{\mu_{\tau}}{\mu_{\text{global}}} \approx \frac{1/|V_{\tau}|}{1/|V|} = \frac{|V|}{|V_{\tau}|}
\end{equation}
Since $|V| \gg |V_{\tau}|$, it follows that $A \gg 1$.
This proves that the Typed-Centrality score is mathematically guaranteed to be orders of magnitude larger than the global score, allowing to clearly distinguish important nodes ("Hubs") from unimportant ones.
\end{proof}

Standard GraphRAG operates in the domain of $G$, where valid signals are weak ($Score \approx 0$). Our approach operates in $G_{\tau}$, where signals are strong ($Score \gg 0$).
By forcing the LLM to select paths based on amplified $C(v, G_{\tau})$ scores, we effectively filter out topologically irrelevant nodes that act as "distractors" in the global graph. Therefore, the consequence of Theorem \ref{theorem:restoration} is a better context for the GraphRAG.

\section{Background: Centrality Measures for Knowledge Graphs}
\label{sec:background}

In graph analysis, centrality measures quantify the structural importance of nodes. Below, we define the set of centralities used in this work \cite{r11}.

\subsection{Degree Centrality}
Degree centrality captures the number of edges incident to a node. For node \(v \in V\) in graph \(G=(V,E)\):
\begin{equation}
    C_{\text{degree}}(v, G) = \deg(v) = |\{u \in V : (v,u) \in E\}|
\end{equation}

\subsection{Betweenness Centrality}
Betweenness centrality quantifies the extent to which a node lies on shortest paths between other nodes:
\begin{equation}
    C_{\text{bet}}(v, G) = \sum_{s \neq v \neq t} \frac{\sigma_{st}(v)}{\sigma_{st}}
\end{equation}
where \(\sigma_{st}\) is the number of shortest paths between nodes \(s\) and \(t\), and \(\sigma_{st}(v)\) counts how many of those paths pass through \(v\). In KGs, nodes with high betweenness can act as bridges connecting distinct clusters or entity types.

\subsection{Closeness Centrality}
Closeness centrality measures how close a node is to all other nodes:
\begin{equation}
    C_{\text{close}}(v, G) = \frac{1}{\sum_{u \in V} d(v,u)}
\end{equation}
where \(d(v,u)\) is the shortest path distance between \(v\) and \(u\). In type-specific subgraphs, closeness highlights nodes that can efficiently reach other entities of the same type.

\subsection{PageRank}
PageRank extends degree centrality by considering the importance of neighbors recursively. For node \(v\):
\begin{equation}
    PR(v) = \alpha \sum_{u \in \text{In}(v)} \frac{PR(u)}{\deg(u)} + (1-\alpha)
\end{equation}
where \(\alpha \in [0,1]\) is a damping factor and \(\text{In}(v)\) is the set of nodes linking to \(v\). 

\subsection{Eigenvector Centrality}
Eigenvector centrality evaluates node importance based on the importance of its neighbors:
\begin{equation}
    C_{\text{eig}}(v) = \frac{1}{\lambda} \sum_{u \in \text{Neighbors}(v)} C_{\text{eig}}(u)
\end{equation}
where \(\lambda\) is the largest eigenvalue of the adjacency matrix. For KGs, this measure highlights nodes embedded within highly connected and influential substructures.

\section{Experimental Setup}
\label{sec:experimental_setup}

This section describes the experimental setup and evaluation of the Typed-Centrality Grounding framework, focusing on the reasoning comparison between standard GraphRAG with global metrics and type-aware metrics computed via the Symbolic Layer. 

\subsection{Data Layer: POLE Dataset}
\begin{figure}[ht!]
\centering
\includegraphics[width=0.70\linewidth]{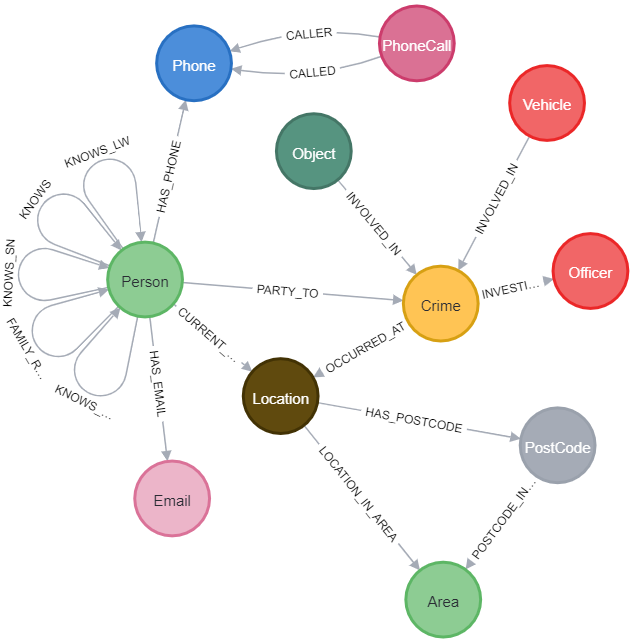}
\caption{POLE Schema.}
\label{fig:pole_schema}
\end{figure}

Experiments use the POLE (Person, Object, Location, Event) dataset, derived from open crime data for the Greater Manchester area\footnote{\url{https://www.data.police.uk/data/}} on Neo4j 1.6.2 (Fig.~\ref{fig:pole_schema}).
The graph schema consists of several distinct node types representing the entities of the POLE model.
The main node class is Person, which is connected to personal identifiers such as Phone (via HAS\_PHONE) and Email (via HAS\_EMAIL).
Social interactions rely on Person self-relationships. Crime nodes define the event context by linking to: people (PARTY\_TO), physical entities (INVOLVED\_IN), officers (INVESTIGATED\_BY) and locations (OCCURRED\_AT).
Geographical data is hierarchical, connecting Location to PostCode and Area. Finally, communication patterns are captured by the PhoneCall node, which links two Phone nodes via CALLER and CALLED edges.

\subsection{Symbolic Layer Evaluation}
To analyze the network's structure and node importance, we calculated the five centrality measures defined above: Degree Centrality, Betweenness Centrality, Closeness Centrality, PageRank and Eigenvector Centrality. Then, we compared Global metrics (baseline) with Typed metrics computed on type-specific subgraphs. 

\begin{figure}[ht!]
\centering
\includegraphics[width=0.75\linewidth]{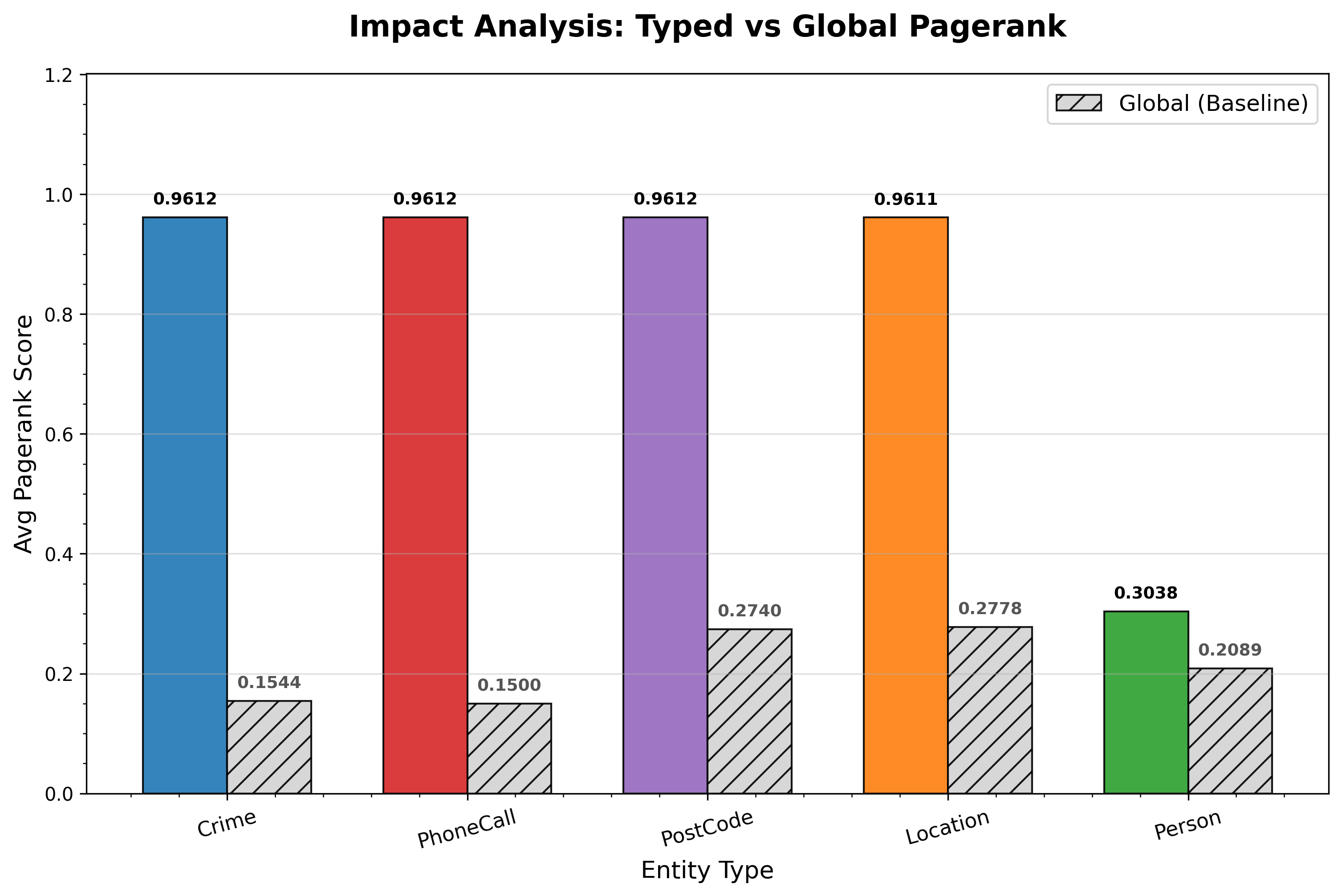}
\caption{Typed PageRank restores signal magnitude compared to the global baseline.}
\label{fig:impact_pagerank}
\end{figure}

\begin{figure}[ht!]
\centering
\includegraphics[width=0.75\linewidth]{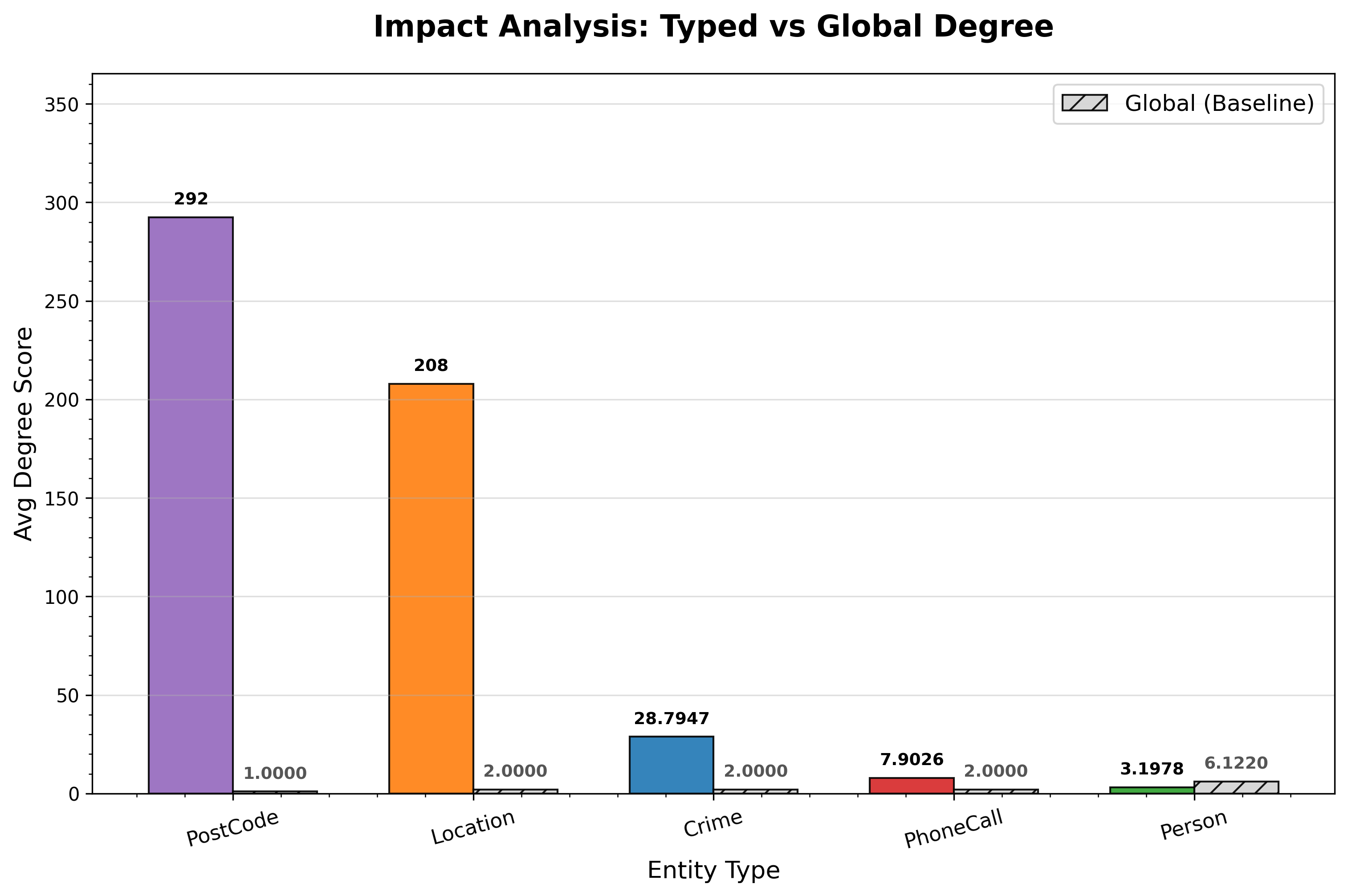}
\caption{Typed Degree reveals local density for PostCode nodes compared to the global baseline.}
\label{fig:impact_degree}
\end{figure}

\begin{figure}[ht!]
\centering
\includegraphics[width=0.75\linewidth]{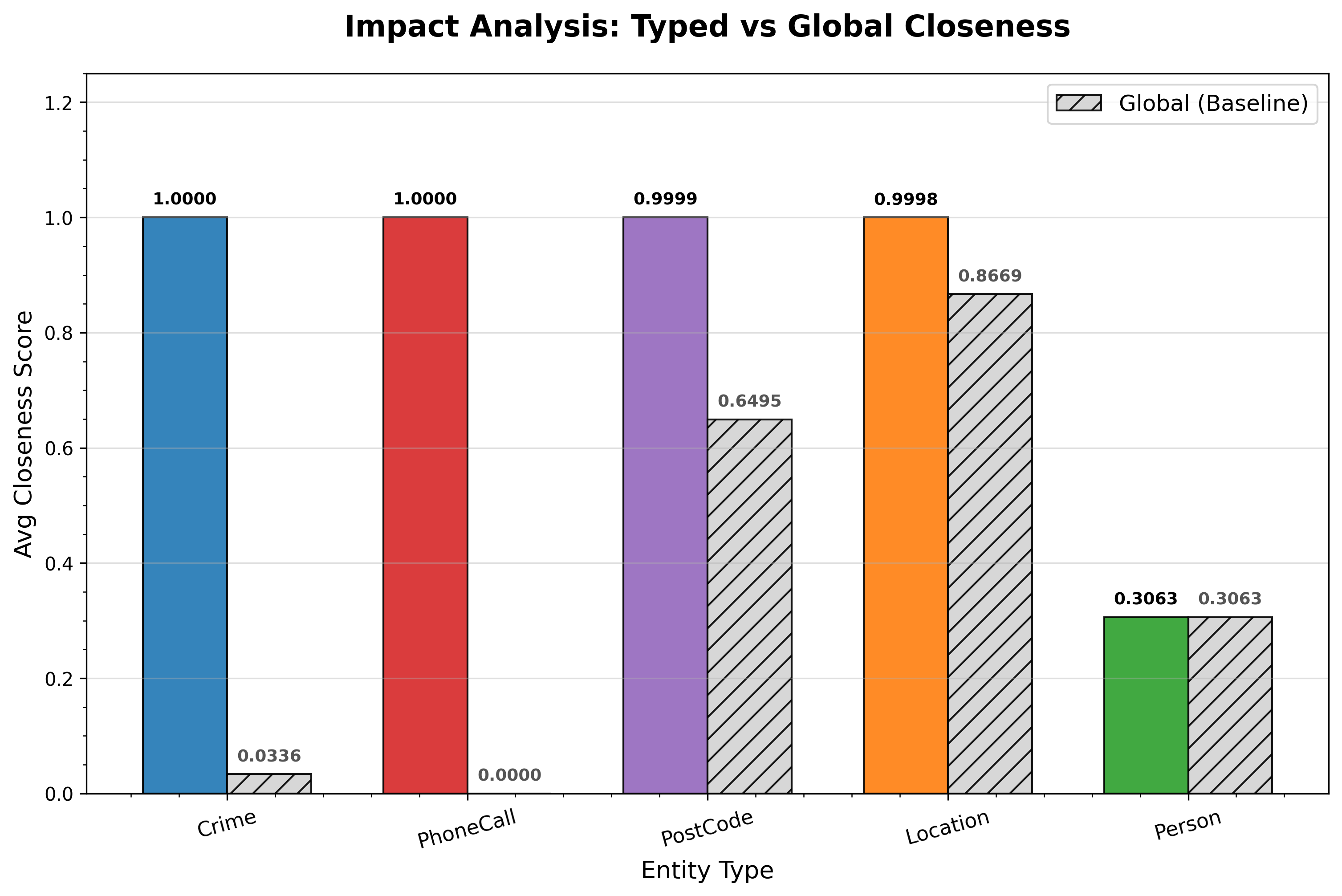}
\caption{Typed Closeness confirms formation of dense hubs.}
\label{fig:impact_closeness}
\end{figure}

Signal magnitude restoration addresses the dilution of global metrics observed in high-cardinality nodes, where values such as the Global PageRank for Crime ($\approx 0.15$) and Global Degree for PostCode ($\approx 1$) are significantly suppressed. By implementing typed metrics, the signal magnitude is successfully restored, reaching approximately $0.96$ for Typed PageRank and $292$ for Typed Degree, reflecting the relevance of these nodes within their specific type-subgraphs (Figs.~\ref{fig:impact_pagerank},~\ref{fig:impact_degree}).
Topological consistency is further validated through Typed Closeness Centrality, which confirms that the projections result in fully connected subgraphs for both Crime and Location entities (Fig.~\ref{fig:impact_closeness}). 

\subsection{GraphRAG and Agentic Layer}
The Similarity Engine fuses textual relevance with Typed-Centrality metrics. Orchestration is handled by Claude Desktop via MCP, enabling the LLM to query the KG and integrate structured signals into reasoning.

\section{Experimental Evaluation}
\label{sec:experimental_eval}

To test our system, we evaluated and compared the responses of two configurations in performing reasoning tasks related to investigative analysis. In particular, the experiments aimed to compare the system’s responses with and without the symbolic layer. Then we wanted to reconstruct the process that led to the generation of specific responses through the typed and global centrality measures analysis.

\subsection{Experiment 1: Resolving Ranking Blindness}

\begin{figure}[ht!]
\centering
\includegraphics[width=0.8\linewidth]{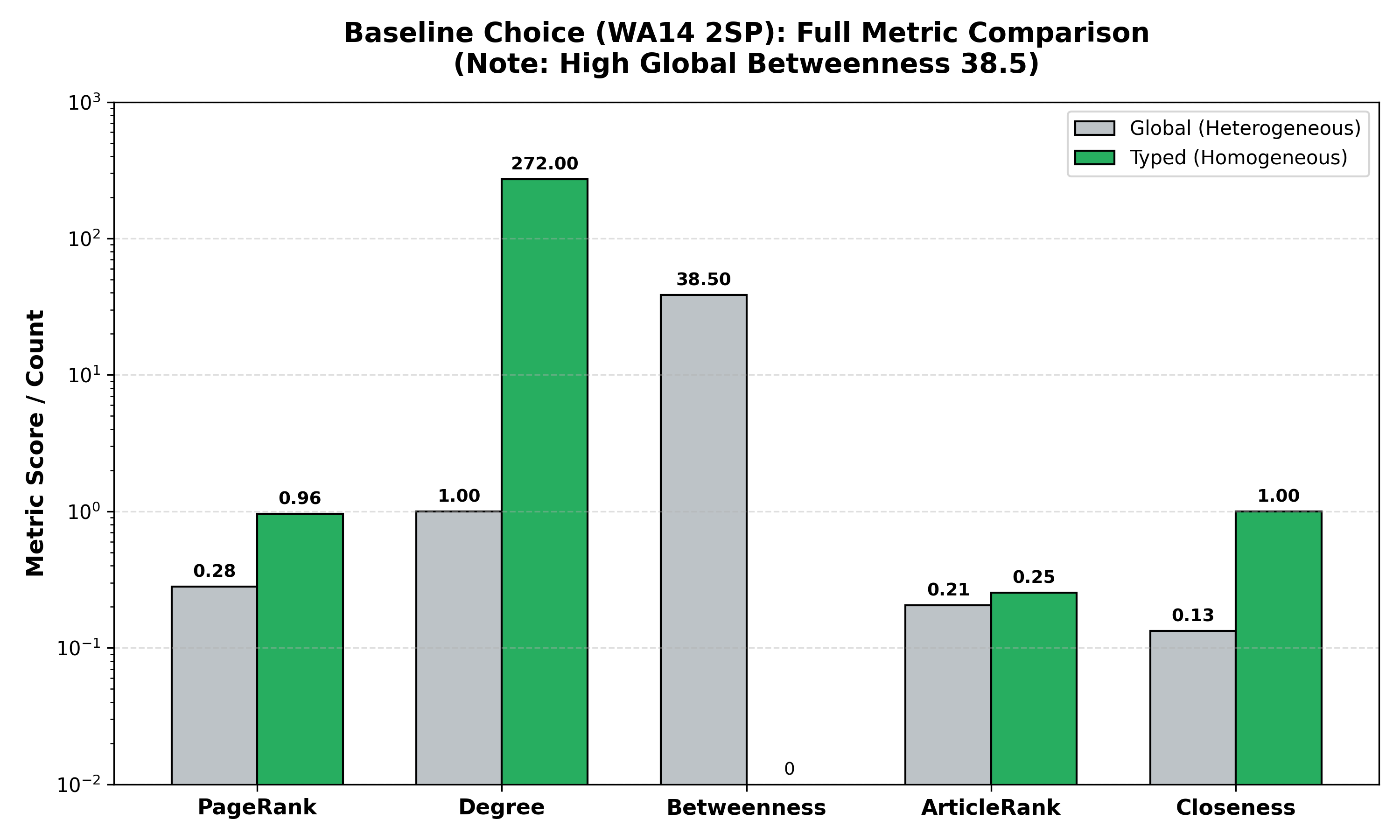}
\caption{Comparison of centrality metrics for PostCode WA14 2SP.}
\label{fig:ranking_blindness_wa14}
\end{figure}

To test type-agnostic ranking limitations, the agent was prompted:

\begin{quote}
\textit{``Using neo4j MCP, identify the `PostCode' entity acting as the most critical structural bridge in the crime network.''}
\end{quote}
The baseline approach, utilizing global metrics, identifies node WA14 2SP due to its high Global Betweenness ($38.5$) (Fig. \ref{fig:ranking_blindness_wa14}). However, this score is primarily driven by connections to numerous unrelated entities across the KGs, such as Person or Location nodes that lack direct relevance to crime events. Consequently, the global metric is amplified, failing to accurately represent the node's role as a bridge within the specific crime subgraph.

\begin{figure}[ht!]
\centering
\includegraphics[width=0.8\linewidth]{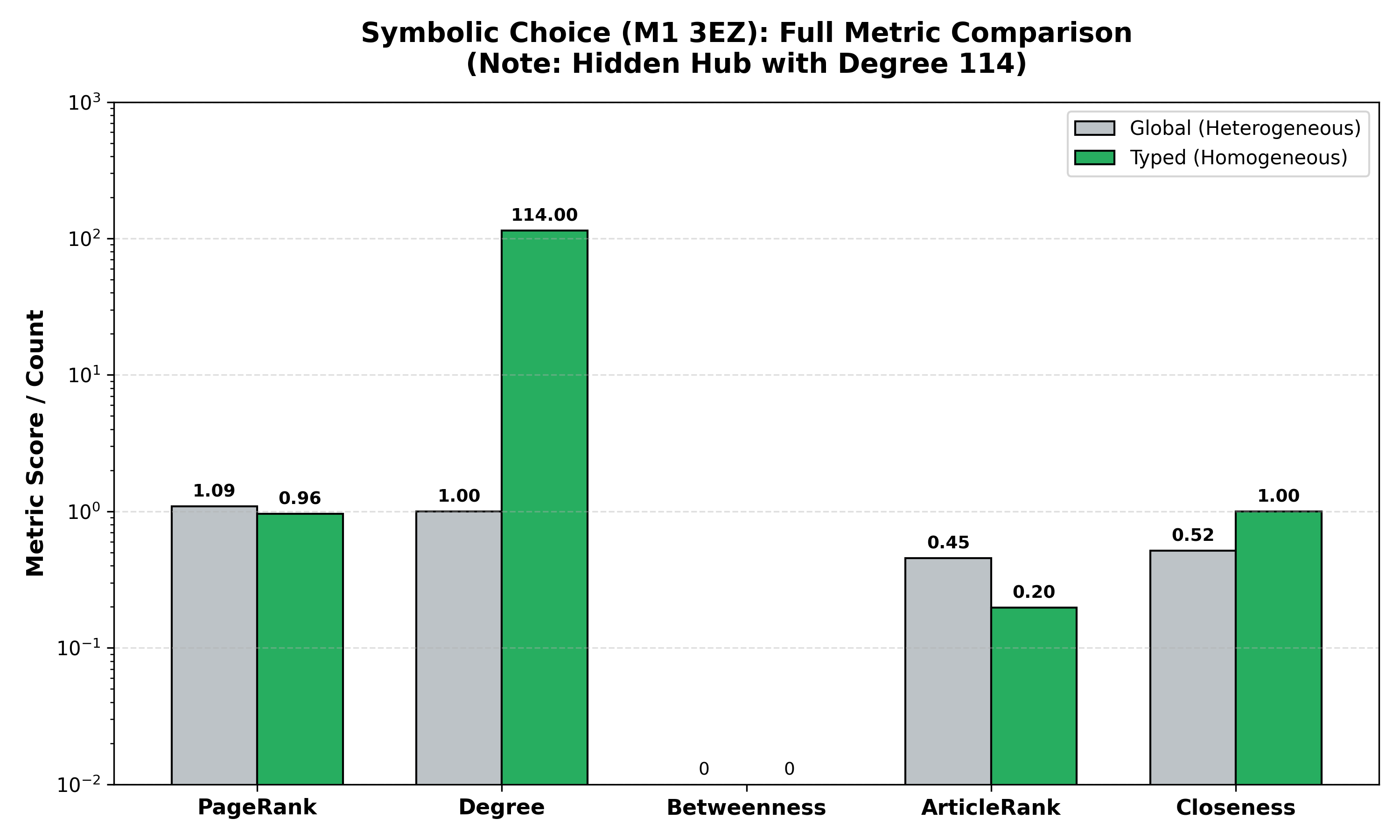}
\caption{Comparison of centrality metrics for PostCode M1 3EZ.}
\label{fig:ranking_blindness}
\end{figure}

In contrast, the typed approach identifies node M1 3EZ as the critical bridge, characterized by a Typed Degree of $114.0$ (Fig. \ref{fig:ranking_blindness}). By computing centrality exclusively within the Location/Crime subgraph, the Symbolic Layer filters out irrelevant nodes from the computation, ensuring that the centrality scores reflect only  real connections. This explains the agent's selection of M1 3EZ, which emerges as a truly central hub within the subgraph connecting crime events.

M1 3EZ also shows a Bridging Efficiency of 144.0 and 288 distinct paths connecting 30 crime events across 7 categories, confirming that its high Typed Degree corresponds to genuine structural importance rather than misinterpretation related to the global graph.

\subsection{Experiment 2: Quantitative Signal Restoration}

\begin{figure}[ht!]
\centering
\includegraphics[width=0.8\linewidth]{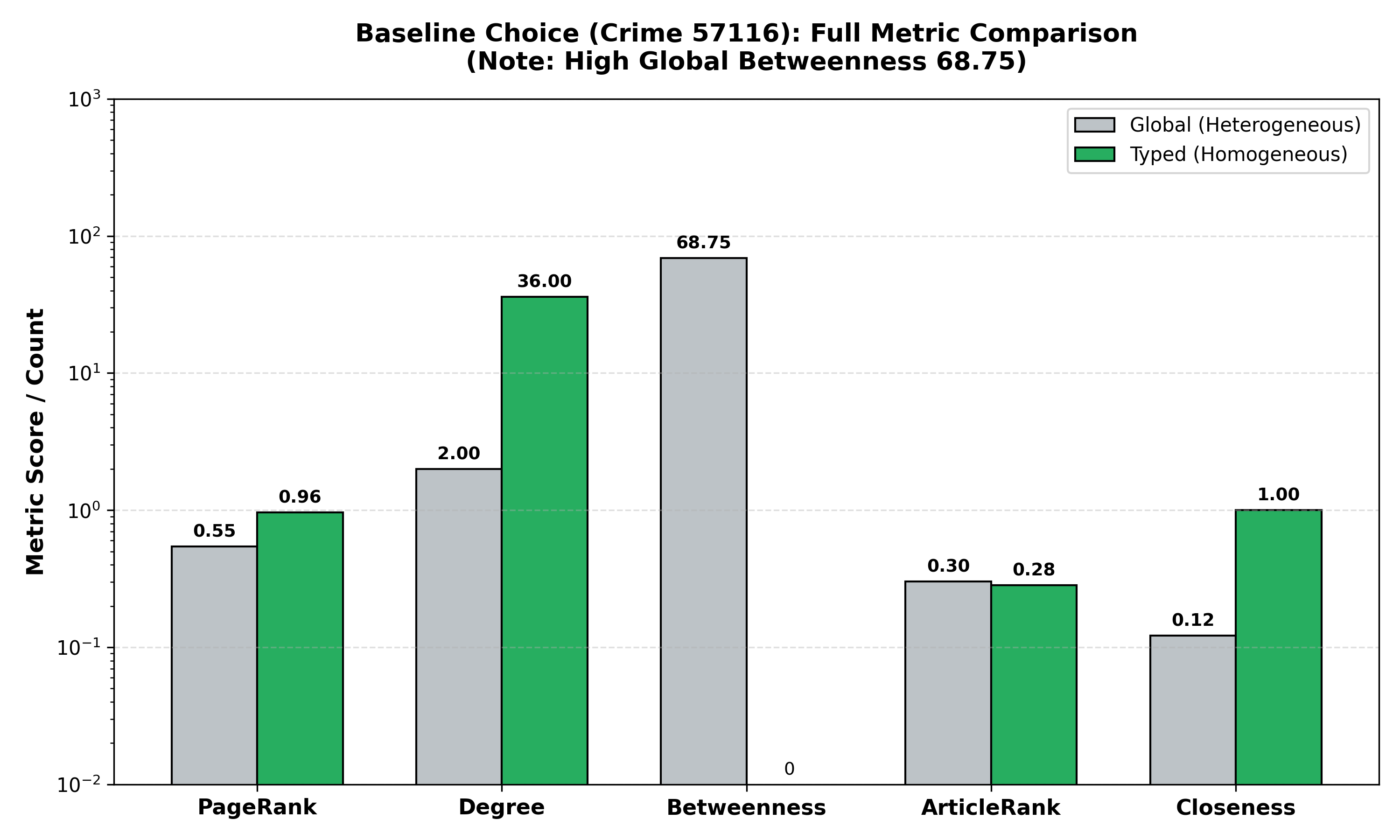}
\caption{Typed-Centrality metrics for Crime Node 57116.}
\label{fig:signal_restoration_57116}
\end{figure}

The second experiment assesses whether type-aware centrality restores meaningful signal magnitude for high-cardinality nodes. The agent was prompted:

\begin{quote}
\textit{``Using neo4j MCP, identify the most influential Crime node based on the importance of its connections within the network structure.''}
\end{quote}
The baseline approach, relying on global metrics, ranks node 57116 as the most significant due to its high Global Betweenness ($68.75$) (Fig. \ref{fig:signal_restoration_57116}). However, this value is artificially inflated by indirect connections to non-crime entities, a factor that dilutes the actual relevance of the node's interactions within the Crime subgraph. This difference is evidenced by its Typed Degree of only $36$, which indicates that the node's influence is significantly less pronounced when restricted to crime-specific interactions.

\begin{figure}[ht!]
\centering
\includegraphics[width=0.8\linewidth]{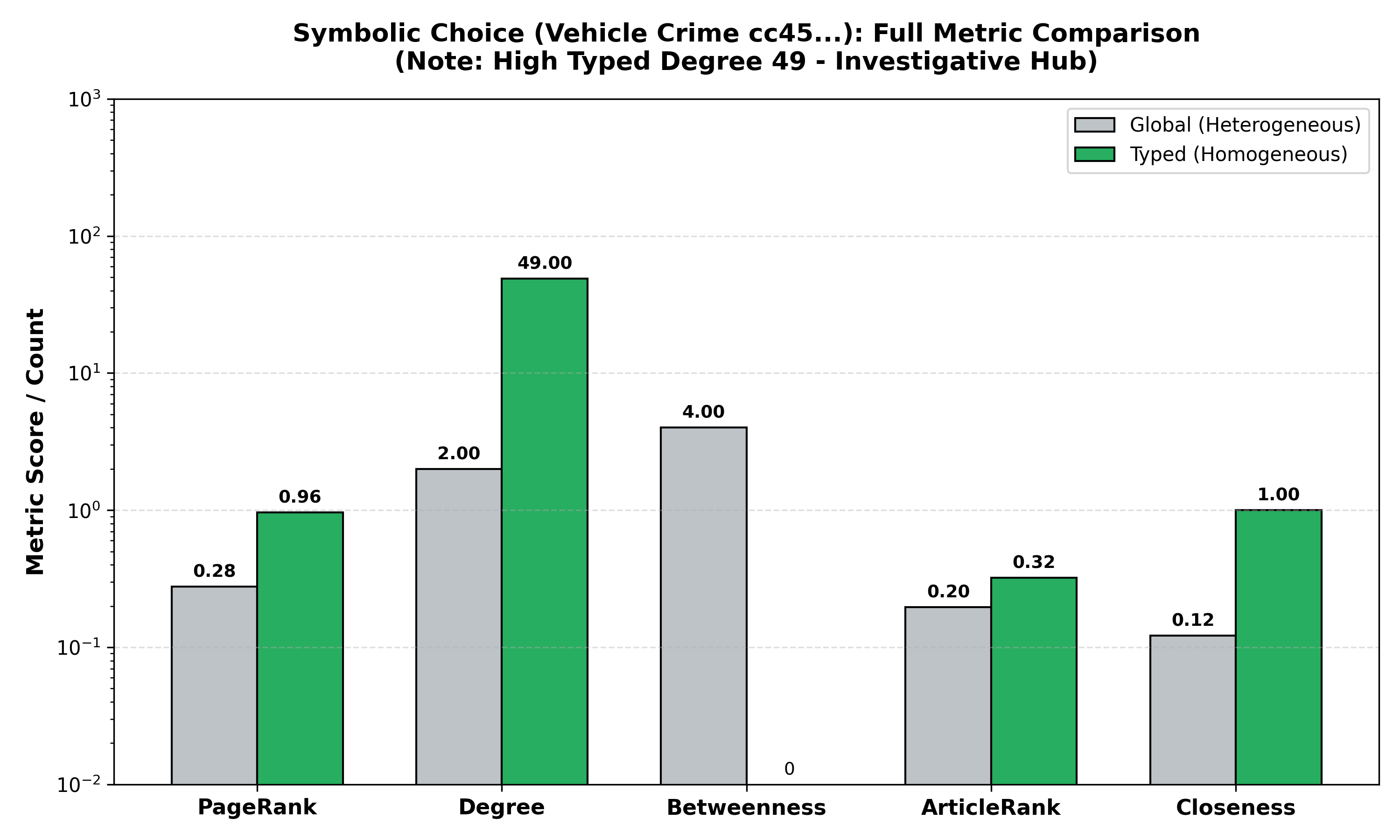}
\caption{Typed-Centrality metrics for Crime Node cc45.}
\label{fig:signal_restoration}
\end{figure}

In contrast, the typed approach identifies node cc45 (Fig. \ref{fig:signal_restoration}) as the top-ranking entity, exhibiting a Typed PageRank of $0.9612$ and a Typed Degree of $49.0$, despite having much lower global counterparts ($0.2775$ and $2.0$, respectively). By computing centrality specifically within the Crime-only projection, the Symbolic Layer restores the relative importance of nodes within their relevant context. This enables the LLM to base its reasoning on structural importance that is aligned with a better model representation, effectively filtering out noise from unrelated subgraphs.

\section{Results Discussion}
\label{sec:results}

The experimental results align with the formal validation of the Typed-Centrality Grounding framework. The model validation demonstrated that global centrality metrics in KGs are affected by signal dilution (Lemma~\ref{lemma:dilution}). In the first experiment, this effect is evident, as the baseline using global metrics selected node WA14 2SP due to high Global Betweenness, even though its connections were mostly outside the relevant Location/Crime subgraph. The Typed approach, which computes centrality within type-specific subgraphs, correctly identified M1 3EZ as the critical bridge. By working on a subgraph, the signal of relevant nodes is amplified relative to irrelevant nodes (Theorem~\ref{theorem:restoration}), allowing the agent to prioritize structurally significant entities. 

In the second experiment, global metrics underestimated the influence of nodes in high-cardinality classes, such as crimes connected to many persons and locations. Node cc45 with Global PageRank of 0.2775 increased to 0.9612 in the Typed projection, and its Degree increased from 2.0 to 49.0. This confirms that the Symbolic Layer restores the magnitude of centrality scores, highlighting nodes that are important within the relevant subgraph.

Overall, the Symbolic Layer improves node selection and signal clarity without modifying the underlying graph data. By filtering out irrelevant nodes and focusing on type-aware subgraphs, the framework provides interpretable supported guidance for agentic reasoning.

\section{Conclusion and Future Work}
\label{sec:conclusion}
In this paper, we presented a Typed-Centrality Grounding for Agentic GraphRAG, a framework designed to address the limitations of symbolic-agnostic reasoning in GraphRAG systems. By introducing a Symbolic Layer that computes typed centrality metrics within type-specific subgraphs, the framework adds a deterministic grounding to the context.
Our approach led to an improved selection of structural hubs, a more accurate ranking of critical nodes, and a clearer identification of subgraph topologies. These symbolic signals guide reasoning and nodes can be better traced by the agent via interpretable typed metrics.

In future work, we plan to extend the framework by integrating a neurosymbolic layer that combines typed-centrality measures with graph embeddings. This will allow the agent to use  both explicit structural signals and learned semantic representations, supporting richer reasoning over the observed KGs. Overall, the Typed-Centrality Grounding Framework demonstrates that KGs can serve as active, interpretable guides for agentic reasoning. By combining type-aware centrality with a symbolic layer that is fed into the GraphRAG, the framework can provide symbolic-informed support for complex reasoning.

\section*{Acknowledgments}
Rosario Napoli is a PhD student enrolled in the National PhD in Artificial Intelligence, XL cycle, course on Health and life sciences. This work has been partially funded by the Horizon Europe ``Open source deep learning platform dedicated to Embedded hardware and Europe'' project (Grant Agreement, project 101112268 - NEUROKIT2E)

\bibliographystyle{unsrt}
\bibliography{bib}

\end{document}